\documentclass[runningheads]{llncs}
\usepackage{amsmath}
\usepackage{graphicx}
\usepackage{amsfonts}
\usepackage{mathtools}
\usepackage{xspace}
\usepackage{hyperref}
\usepackage{url}
\usepackage{color}
\newcommand{\R}{\mathbb{R}}
\renewcommand{\P}{\mathbb{P}}
\newcommand{\D}{\mathcal D}
\newcommand{\U}{\mathcal U}
\newcommand{\T}{\mathcal T}
\newcommand{\problog}{ProbLog\xspace}
\newcommand{\dtproblog}{DTProbLog\xspace}
\newcommand{\prolog}{Prolog\xspace}
\DeclareMathOperator{\util}{Util}

\begin{document}

\titlerunning{An Automated Negotiation Literature review}
\title{The current state of automated negotiation theory: a literature review\thanks{This research was sponsored by the U.S. Army Research Laboratory and the U.K. Ministry of Defence under Agreement Number W911NF-16-3-0001. The views and conclusions contained in this document are those of the authors and should not be interpreted as representing the official policies, either expressed or implied, of the U.S. Army Research Laboratory, the U.S. Government, the U.K. Ministry of Defence or the U.K. Government. The U.S. and U.K. Governments are authorized to reproduce and distribute reprints for Government purposes notwithstanding any copyright notation hereon. }}
\author{Sam Vente\inst{1}\orcidID{0000-0002-5485-6571} \and
Angelika Kimmig\inst{1}\orcidID{0000-0002-6742-4057} \and
Alun Preece\inst{1}\orcidID{0000-0003-0349-9057} \and
Federico Cerutti \inst{2}~ \inst{1}\orcidID{0000-0003-0755-0358}}

\authorrunning{S. Vente et. al.}

\institute{Cardiff University, Cardiff, CF10 3AT, UK 
\email{\{VenteDA,KimmigA,PreeceAD,CeruttiF\}@cardiff.ac.uk} \and
Department of Information Engineering, University of Brescia, Italy \email{federico.cerutti@unibs.it}}

\maketitle

\begin{abstract}
Automated negotiation can be an efficient method for resolving conflict and redistributing resources in a coalition setting. Automated negotiation has already seen increased usage in fields such as e-commerce and power distribution in smart girds, and recent advancements in opponent modelling have proven to deliver better outcomes. However, significant barriers to more widespread adoption remain, such as lack of predictable outcome over time and user trust. Additionally, there have been many recent advancements in the field of reasoning about uncertainty, which could help alleviate both those problems. As there is no recent survey on these two fields, and specifically not on their possible intersection we aim to provide such a survey here. 
\end{abstract}

\keywords{Automated negotiation\and Multi-agent systems\and Constraints}

\section{Introduction}

Negotiation is an important method of conflict resolution and resource allocation in multi-agent systems. In this survey I will focus on \textit{bilateral} negotiations, meaning negotiations with only two parties, but many of the techniques will also be applicable to multilateral negotiations. The current state of the art in automated negotiation places a lot of emphasis on \textit{Proposal Based Negotiation} (PBN) \cite{Baarslag2016} which will be explained in more detail in Section \ref{negPrelim}. This means that the only method of communication the protocol allows is either accepting a previous proposal, offering a counter-proposal or terminating the negotiation altogether.

Work has already been done to develop negotiation support systems (NSS) \cite{pocketNegotiator} and verify that they do transfer benefits to the user when used \cite{Han2005,Lin2009,Wu2009}. Since emotion can have big impacts on how people act both inside and outside negotiations, affective computation, that is computation relating to human emotion, could potentially be worthwhile integrating into NSS. In fact, authors such as Broekens, Jonker and Meyer \cite{Broekens2010} assert that integrating useful affective computation into NSS is possible in the short term (several years). Despite this work, however, as identified by Baarslag et. al. in \cite{Baarslag2009}, and by Pommeranz et. al. in  \cite{Pommeranz2012} adoption of autonomous negotiation agents and NSS systems still faces significant challenges. Baarslag et. al.  \cite{Baarslag2009} identify three main challenges to more widespread adoption: 
\begin{enumerate}
    \item Necessity for domain knowledge and difficulty of preference elicitation
    \item Long-term perspective (having non-stationary preferences and performing well over multiple encounters) 
    \item User trust
\end{enumerate}
They identify that the absence of predictable outcome and transparent preference elicitation form a major barrier to overcome, especially to non specialist users. They also mention that there exists a tension between predictability and performance, since if an agent is predictable to its adversary, it becomes exploitable, meaning that an agent should be predictable but only to the user. They mention that this can be alleviated by, amongst others, quantifying uncertainty. Therefore, it is my hypothesis that it is beneficial to focus on incorporating models of computation that can inspire more user trust by providing the following:
\begin{enumerate}
    \item Explanations for the agent's beliefs and actions
    \item Certain guarantees or quantifying the risk of certain outcomes occurring
    \item Assessments of how likely an opponent is to honour the outcome of a negotiation by measuring credible commitment 
\end{enumerate}

I argue that one way to achieve better transparent consequences is through integrating a Reasoning About Uncertainty (RAU) approach into negotiations. RAU is a well known and well-studied problem with a mature body of research written about it \cite{preece2019explainable,10.1117/12.2519945,kaplan2018uncertainty,cerutti2018probabilistic,halpernReasoningUncertainty2017}. RAU also often offers models that are in some way more explainable than many other systems. User trust often not only has to do with the outcome of a certain task but also the wider implications of that outcome. Using methods such as probabilistic reasoning will allow an agent to consider the wider implications and present these to the user. It is, therefore, my belief that integrating an RAU approach into automated negotiation is a good way to achieve the goals outlined above.

\section{Basic Negotiation theory}\label{negTheory}

In this section I will discuss relevant theory of automated negotiation, summarising work from  \cite{Baarslag2016,Baarslag2013a,Carnevale1993,Hu2013,Kakimoto2014,Poole2017,Rahwan2003}
\subsection{Negotiations}\label{negPrelim}
\begin{figure}
    \centering
    \includegraphics[width=0.9\linewidth]{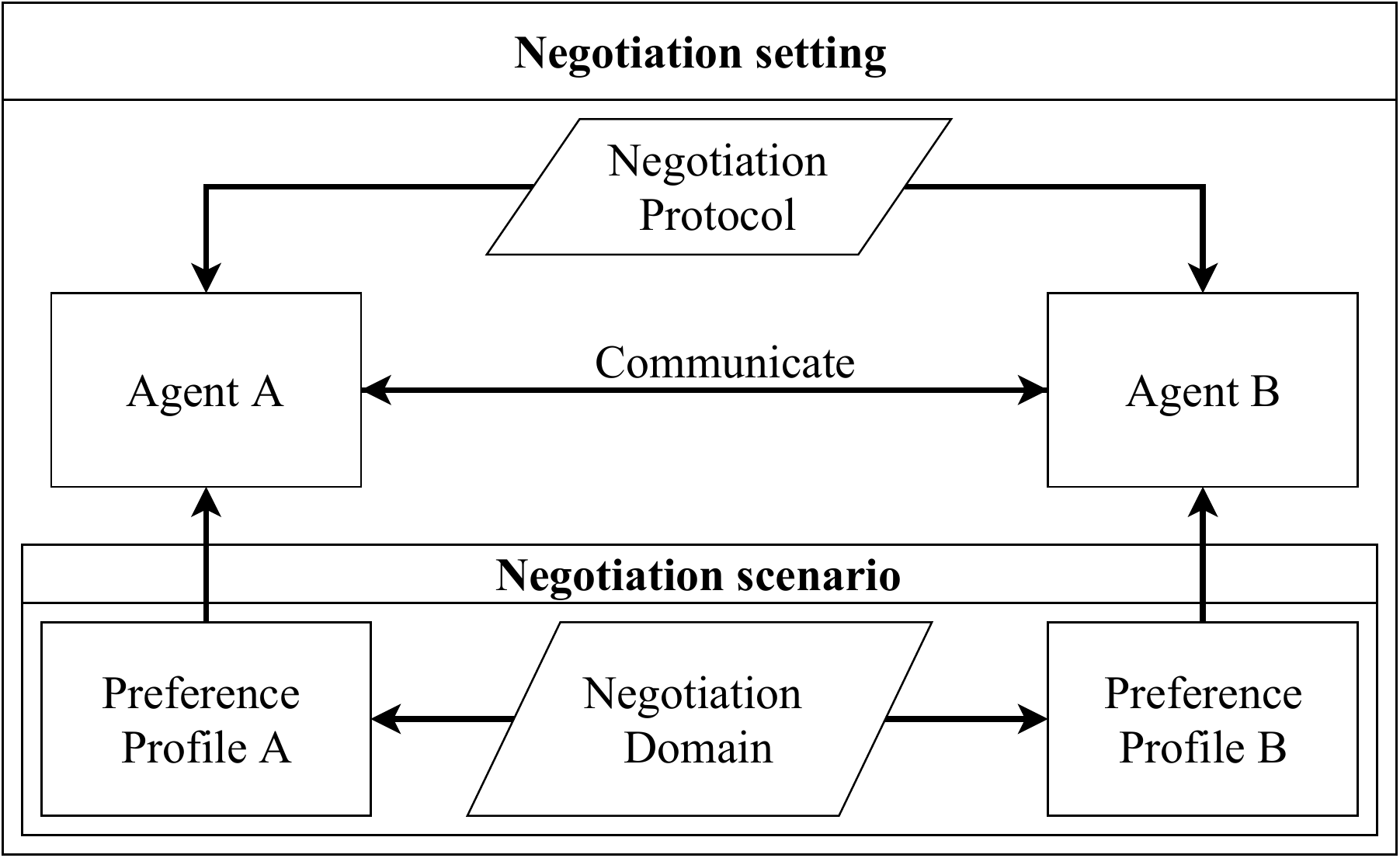}
    \caption{An illustration of the elements defining an automated bilateral negotiation. Reproduced from \cite{Baarslag2016}}
    \label{negOverview}
\end{figure}

A high-concept illustration of the components of an automated negotiation is depicted in Figure \ref{negOverview}. The \textit{negotiation setting} describes the entire negotiation. Its main components are the agents, the \textit{negotiation protocol} and the \textit{negotiation scenario}. We will discuss these elements in more detail below.

\subsubsection{PBN Protocols}
 The negotiation protocol details the methods of communication. In this work, it is assumed that both agents adhere to the same protocol. One of the more well-known protocols is the Alternating-Offers Protocol (AOP). In this protocol, the two agents take turns making offers, or bids to each other until one of the agents accepts \cite{Baarslag2016}. When we refer to \textit{Proposal Based Negotiation} (PBN) we refer to negotiations that utilise AOP. While the AOP is specifically focused on bilateral negotiations, there are also versions for multilateral negotiations. Such an example is the Stacked Alternating Offers Protocol (SAOP) \cite{Aydogan2017} where every agent can make a new proposal or choose to accept the most recent proposal. In general, PBN can also refer to negotiations using a multilateral version AOP (such as SAOP, but more exist), but those are not discussed here. 

Of course, different protocols provide different benefits and disadvantages. Therefore it is important and also non-trivial to choose the correct protocol for framing a given problem. In \cite{Marsa-Maestre2014} Marsa-Maestre et. al. provide rules-of-thumb and strategies for matching a beneficial method of modelling the negotiation based on the problem details, including which protocol to use. 

\subsubsection{Negotiation scenarios}\label{negSenario}
The negotiation domain consists of the matters being discussed in the negotiation, meaning that the domain comprises the semantic part of the interaction. Formally we say that the \textit{negotiation domain}, \textit{negotiation space} or \textit{outcome space}, which is usually denoted as $\Omega$, is the product of at least one \textit{issue}. An issue, denoted $\Lambda$, is a countable set whose elements are called \textit{values}. Each issue represents one matter that needs to be agreed upon, and the values represent possible assignments. A \textit{proposal} is a vector $\omega\in\Omega$, representing a possible assignment to the issues being proposed. 

The negotiation domain informs the \textit{preference profiles} of the agents. An ordinal preference profile is a linear preorder $\succeq$, meaning a linear ordering that is not necessarily anti-symmetric, on the outcome space. We say that an agent \textit{weakly prefers} $\omega$ to $\zeta$ if $\omega \succeq \zeta$. If in addition $\zeta \succeq \omega$ holds, we say that the agent is \textit{indifferent} towards the offers, which is written as $\zeta\sim\omega$. If $\omega \succeq \zeta$ holds but $\zeta \succeq \omega$ does not, then we say that $\omega$ is \textit{strictly} or \textit{strongly preferred} to $\zeta$, and write $\omega \succ \zeta$ \cite{Poole2017}.
Finally, a proposal $\rho$ is called \textit{Pareto optimal} if it cannot be improved upon for one agent without making it worse for another. More formally this means that a proposal $\rho$ is Pareto optimal if, whenever an agent prefers $\zeta$ to $\rho$ then the other agent prefers $\rho$ to $\zeta$. This is a measure of the efficiency of an agreed proposal, because if a proposal is not Pareto optimal, there exists an offer that all parties prefer or are indifferent to. The piece-wise linear curve connecting all Pareto optimal solutions is called the \textit{Pareto frontier}. An agent may also specify a \textit{reservation value}, which is the least favourable proposal that is still preferable to ending the negotiation without an agreement \cite{Baarslag2016}.

Although an ordinal preference profile can be formulated by listing all of the ordinal relationships, for large negotiation spaces this quickly becomes unwieldy. A more common way of formulating a preference profile is by use of a \textit{utility function}. This is a function $u: \Omega \to \R$, which assigns a numeric value to each outcome which represents how much the agent appreciates that outcome. Every utility function $u$ induces a preference profile via the following relationship \begin{equation}\omega \succeq \zeta \iff u(\omega) \geq u(\zeta).\end{equation} Such preference profiles are called \textit{cardinal} preference profiles. Sometimes additional assumptions are made about utility functions to help alleviate computational complexity but this almost always comes at the cost of expressiveness. One common way of doing this is by using \textit{linear additive functions} for which it holds that for any $\omega \in \Omega$ where $\Omega$ consists of $n$ issues, the following equation holds \begin{equation}u(\omega) = \sum_{i=1}^n w_i e_i(\omega_i)\end{equation}\cite{Baarslag2016}. Here $w_i$ are normalised weights representing the relative importance of each issue. 

\subsubsection{Time} In the real world, negotiation is often very costly both because the issues debated might be time sensitive (resources decaying for example) and because the act of negotiating itself requires resources. Therefore it is often desirable to take into consideration the time needed to negotiate \cite{Carnevale1993}. One common way to model this is by using a \textit{time discounted} utility. For a certain discount factor $\delta \in (0,1]$ and $t \in [0,1]$ the normalised time this is defined as \begin{equation}
    u^\delta(\omega) = u (\omega) \cdot \delta^t.
\end{equation} Here a small $\delta$ represents high time pressure and $\delta=1$ represents no time-pressure. \cite{Baarslag2016}

\subsection{Success metrics}\label{metrics}
\subsubsection{Measuring negotiation outcomes}
There are a variety of metrics commonly used to assess the outcome of a negotiation:

\begin{itemize}
    \item \textit{Average utility} This is by far the most common metric used to assess the efficiency of a negotiation agent. The average utility an agent can achieve is indicative of its performance. This metric does have its limitations however, since it can provide only a metric relative to a certain utility function, and cannot provide a more general notion of performance. For example, if one agent uses a utility function $u$ and another uses the same strategy but uses the utility function $w(\omega) = 10 \cdot u(\omega)$ they will make the same decisions but the agent using $w$ will in general appear to perform much better.  
    \item \textit{Fairness of outcome} In some negotiation settings, such as cooperation settings, or setting where repeated interactions are likely, the fairness of an outcome can be an important factor. Often this is characterised as the average distance to solutions that satisfy fairness characteristics such as the Nash solution, or the Kalai-Smorodinsky solution. Here the Nash solution is defined as the bid which maximises the product of the utilities \cite{Baarslag2016}: \begin{equation}
        \omega_{Nash} = \max_{\omega \in \Omega} u_a(\omega) \cdot u_b(\omega)
    \end{equation} and the Kalai-Smorodinsky solution is defined as the Pareto optimal bid $\omega$ for which the following equation holds: \begin{equation}
        \frac {u_{max_a}} {u_{max_b}} = \frac{u_a(\omega)}{u_b(\omega)}
    \end{equation} meaning that each agent gets a utility proportional to their maximal possible utility. The main difference between these measures is that Nash equilibria are stable under removing irrelevant options from the domain, while the Kalai-Smorodinsky solution guarantees that if there are more resources to divide, then agents won't have their utility decreased by claiming them \cite{Kalai1975}.
    \item \textit{Distance to Pareto frontier} As I mentioned in Section \ref{negSenario}, a Pareto bid is in some ways maximally efficient because if a bid is not Pareto optimal, by definition there exists a bid that is better for at least one agent and nobody will have to concede. Therefore the average distance to the nearest Pareto bid is often used as a measure of the efficiency of an outcome. One problem with analysing Pareto optimally is that it requires perfect information. However, preference strategies and thus Pareto optimal bids can be estimated using machine learning, which will be explored in more detail in section \ref{negML}. Calculating the Pareto frontier can be costly, especially since it suffers from the curse of dimensionality. Genetic algorithms are often used to approximate it such as in \cite{Kakimoto2014} but exact algorithms have also been proposed \cite{Hu2013}.  
    \item \textit{Joint utility} Another measure of fairness that is sometimes used is the joint utility since this in some ways can compensate for exploitation. Many authors simply use the sum of the utilities. However, this can have sensitivity issues since it has trouble detecting unfair outcomes if the scale of the utilities greatly differs. Another possibility is to take the geometric mean of the utilities, as shown in equation \ref{geomMean}, which suffers less from the sensitivity problem. 
    \begin{equation}\label{geomMean}
        u_{joint}(\omega) = \sqrt{u_a(\omega) \cdot u_b(\omega)}
    \end{equation}
    \item \textit{Robustness} Especially when assessing the possible autonomy of an agent and its long term viability, it is often important to consider its robustness. Robustness is a measure of how easily the agent can be exploited by an adversary switching tactics. Highly specialised agents can produce very good outcomes but if they have no means of adapting, they can easily become exploitable by agents who have knowledge of the strategy \cite{Baarslag2013a}. Often robustness is measured using some form of Empirical Game-Theoretic analysis (EGT), sometimes also called Evolutionary Game Theoretic analysis. Examples of works using this kind of analysis include \cite{Baarslag2013a} and \cite{Chen2014}. EGT attempts to provide insight into the robustness of agents by simulating many negotiations among different agents while letting agents change strategies based on the outcomes of the previous round of simulations. Robustness can then be measured by observing the evolution of the distributions of the strategies over time. Work has also been done recently by Tuyls et. al. \cite{Tuyls2018} to prove theoretical guarantees on an EGT analysis technique called meta-game analysis. This technique models higher level strategies using EGT instead of the atomic actions. 
    \item \textit{Time until agreement} As stated earlier, many negotiations have some kind of time pressure. In those scenarios it can be useful to measure the time until an agreement is reached, if at all. Depending on the scenario, this time can be measured in real time, or in the number of messages exchanged, which might be more important in certain low bandwidth scenarios. One disadvantage of this method is that it does not take the quality of the outcome of the negotiation into account.
\end{itemize}

\subsection{Argumentation based negotiation}
In \cite{Rahwan2003} Rahwan et. al. mention that models of PBNs tend to have several disadvantages. For example, usually it is assumed that the preference profile is static and completely known prior to the interaction. This presents problems when agents have to negotiate more complex situations such as trade union negotiations where the value of an outcome is not only unclear, but debatable. Another example is that sometimes agents can fail to reach an agreement because of incorrect assumptions, or the fact that some of the assumptions of perfect rationality cannot always be guaranteed for agents due to matters such as computational resource constraints. In an attempt to address these problems work has been done to create protocols for negotiation that allow more than just proposals to be exchanged. This type of negotiation are referred to as Argumentation Based Negotiations (ABN). This does, however, bring with it some unique challenges. In addition to the challenges any agent should overcome, such as when to agree, when to walk away, and generate bids. Rahwan et. al. \cite{Rahwan2003} identify several other challenges which argumentation based agents must also address:  
\begin{itemize}
    \item \textit{Message interpretation:} In contrast to PBN a message in an ABN may or may not contain, amongst others, a proposal, a rejection, an acceptation, an argumentation for or against a fact, an assertion or anything else that is allowed by the protocol specification used. All these possibilities need to be parsed and differentiated so that they can be dealt with appropriately. When I say message interpretation, that usually refers to the parsing of the low-level symbols into a more semantic representation of the message. For example, imagine that agents A and B are in an ABN. When A receives a message containing the literal string \texttt{"assert(isAlly,A,B)"}, then message interpretation refers to the process of converting this string into an internal representation of the fact that B is asserting that A and B are allies (whatever that may mean in context). This is more challenging than in the PBN case since all of the possibilities mentioned above might have their own associated symbols and/or grammar. For example, consider the difference between how the (arbitrary) messages \texttt{"assert(isAlly,A,B)"}, and \texttt{"propose(yes,yes,no,4,3,no)"} should be parsed. The first is an assertion that can be accepted or rejected and the second is a proposal that has to be considered in a very different way based on its utility. Several options for communication standards are explored in \cite{Casali2016}.
    \item \textit{Message evaluation:} In ABNs evaluating messages is no longer only limited to calculating the utility of an offer, but must involve some form of reasoning. For example take the example of \texttt{"assert(isAlly,A,B)"}. First of all the agent must decide whether to accept or reject this assertion and then the agent must consider the implications and how it values those. 
    \item \textit{Message generation:} In addition to generating new proposals, an agent must now also be able to generate new arguments and decide when it is appropriate to do so.
    \item \textit{Belief revision:} Although not technically necessary, a belief state is integral to many forms of argumentation\cite{Amgoud2012,Rahwan2003}. In addition to that, several authors also note that incorporating belief revision into an agent has benefits, such as being able to avoid non-agreement due to incorrect assumptions \cite{Casali2016,Pilotti2015}. This opens up a potential way of modelling credible commitment or assessing how likely it is an agent can and will make good on its agreements. This is an avenue which is not explicitly addressed in the current literature.
\end{itemize}

\subsubsection{ABN Protocols}
A defining trait of an ABN is that the protocol in some way allows agents to express the reasoning behind a bid or rejection. In \cite{Rahwan2003}, Rahwan et. al. mentions that in addition to the language ABN agents use to communicate, they also often need a \textit{domain language} to refer to concepts of the environment which are relevant to the negotiation.  One common standard protocol is the FIPA ACL protocol \cite{El-Sisi2012,FIPA,FIPAACL}, which is an architecture specification and an associated protocol specification that attempts to promote inter-operability between negotiation agents. FIPA ACL does not specify any domain language, leaving the interpretation of the messages to the agent. 

Some work has also been done by Luo et. al. in \cite{Luo2003a}. They propose a method of negotiation based on the AOP protocol, but also develop mechanics to allow agents to express fuzzy constraints to each other, which they can use to find Pareto optimal solutions.

\subsection{Negotiation and machine learning} \label{negML}
\subsubsection{Forms of opponent modelling}
Opponent modelling can aid an agent to achieve better outcomes in a number of ways, but the most common ones are being able to avoid exploitation by adapting to the opponent, minimising the cost of negotiations and being able to reach win-win arguments. While the modelling techniques mentioned below can be applied to any negotiation, their viability depends more on details of the encounter such as the availability of relevant data, or the representation of the preference profiles that agents use. 

\begin{itemize}
    \item \textit{Learning the acceptance strategy} Having information about the acceptance strategy of the opponent has a huge strategical advantage since it allows an agent to look for the bid that the agent itself prefers the most which is still acceptable to the opponent. However, this is often kept secret to avoid exploitation. A popular way of learning the acceptation strategy is estimating the \textit{acceptance probability}, which is the probability that the agent will accept any given proposal. Sometimes it can also be beneficial to model the opponent's deadline since that can reveal whether stall-tactics can be effective or not \cite{Baarslag2016}. 
    \item \textit{Learning the bidding strategy} Having information about an opponent's bidding strategy and adjusting one's own preference information accordingly can be beneficial. For example, demanding higher concessions from agents that are known to concede easily can lead to better outcomes. However, this is a challenging task, especially since an agent's strategy may not be static over time. Time series forecasting is sometimes used to predict the future behaviour of opponents \cite{Baarslag2016}.
\end{itemize}

Conversely, the aim of using an opponent model is usually to increase performance, therefore it is also very popular to measure the difference in performance when using the model. For this any of the metrics outlined in Section \ref{metrics} can be used.

\section{Reasoning about uncertainty}\label{RAU}
 It is my belief that RAU provides methods of computation that can satisfy the goals I have set out in the introduction of this piece. Therefore I will first give a brief overview of the theory of probabilistic reasoning I intend to employ to achieve those goals. 
 \newpage

\subsection{\dtproblog}\label{DTPLPrelim}

\subsubsection{First order logic} While a basic understanding of first order logic is assumed, I will give a very brief overview of the used terminology.
\begin{itemize}
    \item A \textit{constant} is a symbol referring to a specific object such as 2 or \texttt{bart}
    \item A \textit{variable} is a symbol referring to a wider variety of objects such as \texttt{X} or \texttt{Person} Here I use the convention that logical constants start with a lower case letter and logical variables start with an upper case letter.
    \item A \textit{functional expression} is a functional constant, together with $n$ terms called its \textit{arguments} enclosed in parenthesis and separated by commas. Here $n$ is called the \textit{arity} of the function. For example if $f$ is a functional constant of arity 2 and \texttt{a} and \texttt{Y} are terms (to be defined below), then \texttt{f(a,Y)} is a functional expression. 
    \item A \textit{term} is a variable, a constant or a functional expression.
    \item A \textit{literal} is an atomic logical sentence, or its negation, such as $p$ or $\neg q$.
    \item  A \textit{clause} is a disjunction of literals, such as $p \wedge \neg q$.
    \item A \textit{predicate} is an assertion such as \texttt{isFatherOf(homer,bart)}.
    \item  An \textit{atom} is a sentence consisting of a single predicate. 
\end{itemize}  A logical expression is called \textit{ground} if it contains no variables, i.e.  \\\texttt{isFatherOf(X,bart)} is not ground while  \texttt{isFatherOf(homer,bart)} is. 

A set of clauses is often referred to as either a \textit{program} or a \textit{knowledge base} ($KB$). The set of all ground atoms occurring in a knowledge base $KB$ is its \textit{Herbrand base}. A \textit{Herbrand interpretation} is a mapping from a Herbrand base onto the set of truth values $\{true,false\}$. Whenever a Herbrand interpretation maps an atom or clause to $true$ this interpretation is called a \textit{model} for that atom or clause. An interpretation is a model for a knowledge base if it is also a model for every clause in that knowledge base  \cite{Flach1994}.  For a knowledge base $KB$ and clause $\varphi$ we say that $KB$ \textit{logically entails} or \textit{ models } $\varphi$ if every model for $KB$ is also a model for $\varphi$, which is written as $KB \models \varphi$.  Furthermore, if $R$ is a set of inference rules, and $\varphi$ can be deduced using the rules in $R$ from a set of clauses called $\Delta$, then we say that $\varphi$ is \textit{provable} in $\Delta$, and we write $\Delta \vdash_R \varphi$ or often $\Delta \vdash \varphi$ if $R$ is understood. If $\Delta \models \varphi$ holds whenever $\Delta \vdash_R \varphi$ holds $R$ is called \textit{sound}, which is to say that everything that is provable from a set is also logically entailed by it. If the reverse always holds, $R$ is called \textit{complete}.  

A \textit{substitution} $\theta$ is a mapping from variables to terms. For example, let  $\theta = \{\texttt{X} \to \texttt{homer}, \texttt{Y}\to \texttt{bart}\}$ and $\varphi = \texttt{isFatherOf(X,Y)}$ then we can apply $\theta$ to $\varphi$ which is denoted as $\varphi\theta$ which results in \texttt{isFatherOf(homer,bart)}. Note that in this example $\varphi\theta$ is ground. When this is the case, we call $\theta$ a \textit{grounding} of $\varphi$ \cite{Genesereth2013}.

\subsubsection{\prolog}
As mentioned in \cite{Russell2009} \prolog is one of the most widely used logic programming languages. A \prolog program is a set of \textit{definite} or \textit{Horn clauses}, meaning sentences or clauses in first-order logic containing at most one positive literal, such as $u \vee \neg p \vee \neg q$, which can be rewritten as $(p \wedge q) \rightarrow u$. Here $u$ is called the \textit{head} of the clause and $ p \wedge q$ is called the \textit{body}. In  \prolog this is written as $$\mathtt{u :- p, q.}$$ This can be read as ``$u$ \textbf{if} $p$ \textbf{and} $q$".  As mentioned in \cite{Flach1994} \prolog uses what is called Selective Linear Definite clause resolution or SLD resolution, which is both sound and complete for Horn clauses. However, it should be noted that Prolog uses a depth-first search manner to search for proofs, and may therefore not terminate for certain programs. 

In the context of \prolog, a \textit{fact} is a clause with an empty body. A \textit{rule} is a clause with a non-empty body. 

\subsubsection{\problog}
ProbLog \cite{Raedt2007} is a probabilistic extension of Prolog. We will assume basic knowledge of Prolog in this work. For a more detailed instruction see for example \cite{Flach1994}. Where in pure \prolog one is commonly interested in whether a given query fails or succeeds, in \problog one is focused on computing the probability that a query succeeds in a given ProbLog program. A \problog program $\mathcal T = \mathcal F \cup \mathcal{BK}$ consists of a set of labelled probabilistic facts $\mathcal F$, and a set of definite clauses $\mathcal{BK}$ called the background knowledge. Here, $\mathcal F$ is of the form $\mathcal F = \{p_1::f_1,\ldots,p_n::f_n\}$ where $f_i\theta$  is true with probability $p_i$ for all substitutions $\theta$ grounding $f_i$. The probabilities of all the facts are assumed to be independent. Furthermore, let $\mathcal F \Theta$ denote the set of all possible ground instances of the facts in $\mathcal F$ and let $\mathcal F_L \subset \mathcal F \Theta$. Then A ProbLog program defines a probability distribution of all the sub \prolog programs $L  = \mathcal F_L \cup \mathcal{BK}$ via the following identity: \begin{equation} \label{programDistribution}
    \mathbb{P}(L|\mathcal T) = \prod_{f_i \in \mathcal F_L} p_i \prod_{f_i\in \mathcal F \Theta \setminus \mathcal F_L} (1-p_i)
\end{equation} 

From that we can determine the \textit{success probability} of a given query $q$ using the following identities:
\begin{equation}\label{qGivenL}
    \P(q|L) = \begin{cases}
    1 & \exists \theta : L \models q\theta \\
    0 & \text{otherwise}
    \end{cases}
\end{equation}

\begin{equation}\label{qAndLGivenT}
    \P(q,L|T) =\P(q|L)\cdot \P(L|T) 
\end{equation}

\begin{equation}\label{qGivenT}
    \P(q|T) = \sum_{M\subseteq L_T} \P(q,M|T)
\end{equation}
So the probability of a query given a program is the probability of the query having a proof in a randomly sampled subprogram according to the distribution.

\subsubsection{\dtproblog}
DTProbLog \cite{Broeck2010} is an extension of \problog that allows for strategic planning, given some quantified uncertainty by maximising the expected utility of some actions.

A \dtproblog program is of the form $\mathcal T = L_T \cup \mathcal D \cup \mathcal U$, where $L_T$ is a \problog program, $\mathcal D$ is the set of \textit{decision facts} and $\U$ is the set of \textit{utility attributes}. Decision facts are of the form $?::d$ where $d$ is an atom and $?$ indicates that it is a decision variable for which the program should optimise.  A \textit{strategy} $\sigma$ is a function $\sigma: \D \to [0,1]$ denoting the probabilities with which the agent will assign that literal as true. For a set of decision facts $\D$ and a strategy $\sigma$ the following set is also defined:
$\sigma(\D) = \{\sigma(d)::d | (?::d)\in \D\}$.  
Note that $L_T \cup \sigma(\D)$ is a standard ProbLog program and we will denoted it with $\sigma(\D\T)$.

Furthermore a utility attribute is a statement of the form $u_i \to r_i$ where $u_i$ is a literal and $r_i$ is a numerical reward that will be awarded whenever $u_i$ is satisfied. We then define the utility of a subprogram $L$ as \begin{equation}\label{subProgramUtility}
\util(L) = \sum_{(u_i\to r_i)\in\U} r_i\cdot \P(u_i|L) \end{equation}
 where $\P(u_i|L)$ can be computed according to equation \ref{qGivenL}.
Since a ProbLog program $\T$ defines a probability distribution over the subprograms we can define the expected utility of a program as \begin{equation}\label{expectedUtility}
    \util(\T) = \sum_{(u_i\to r_i)\in\U} r_i\cdot \P(u_i|\T) \end{equation}

This leads us to the definition of the utility of a single utility attribute $a_i$ under a given strategy:
\begin{equation}\label{singleUtility}
    \util(a_i | \sigma,\D\T) = r_i \cdot \P(u_i | \sigma(\D\T)) 
\end{equation}
and the total utility of a strategy \begin{equation}\label{stratUtility}
    \util(\sigma,\D\T) = \util(\sigma(\D\T)) = \sum_{a_i\in \U} \util(a_i|\sigma(\D\T))
\end{equation}
So, in other words, the utility of a strategy is defined as the sum of the expected utilities of all the utility attributes. 

\section{Future Work}

\subsection{An RAU implementation for better predictability of outcome and broader awareness}
As detailed in the introduction Baarslag et. al. identify in \cite{Baarslag2009} that a lack of predictable outcome to the user poses a major challenge to overcome to ensure wider adoption of automated negotiations and NSS.  Using probabilistic reasoning to implement negotiation agents provides a clear and inbuilt way of providing that predictability, and also provides a way to incorporate a broader view of the situation into the workings of an agent instead of only the issues pertaining to the actual negotiation. It has already been seen in  \cite{Mirzayi2018} that context awareness has advantages for negotiation agents. Applications using probabilistic reasoning methods such as ProbLog \cite{Raedt2007} could incorporate information from various sensors, such as demonstrated by \cite{deepcep}, to be incorporated in both bidding strategies and opponent modelling. This could provide potentially huge advantages to agents that are able to apply these concepts well. 

\subsection{Constrained proposal methods}
Some work has already been done in \cite{Luo2003a} by Luo et. al. to incorporate fuzzy constraints into a buyer-seller negotiation setting and found that it ensures that agreements are Pareto optimal. In addition it also minimises the amount of information agents have to disclose, and improves the efficiency of the negotiations over standard AOP settings. It has already been shown by \cite{vente2020} that incorporating very simple constraints into an AOP setting can significantly speed up the negotiations without impacting the outcome. The authors did not test scenarios with time discounted utility functions, more complex or perhaps ``approximate" constraints. If agents were to incorporate more sophisticated methods of discovering constraints, reasoning about them and incorporating possible constraints into bidding strategies, this would convey many opportunities to improve both in terms of explainability and performance. 

 \bibliographystyle{splncs04}
 \bibliography{library.bib}

\end{document}